\documentclass{article}

\PassOptionsToPackage{numbers, compress}{natbib}
\usepackage[preprint]{MLstyle}

\usepackage{graphicx}

\usepackage[utf8]{inputenc} 
\usepackage[T1]{fontenc}    
\usepackage{hyperref}       
\usepackage{url}            
\usepackage{booktabs}       
\usepackage{amsfonts}       
\usepackage{nicefrac}       
\usepackage{microtype}      
\usepackage{float}
\usepackage{enumerate}

\usepackage{xcolor}         
\usepackage{amsmath}
\usepackage{amssymb}
\usepackage{amsthm}
\usepackage{amsmath}
\usepackage{microtype}      
\usepackage{xcolor}         

\usepackage{subcaption}
\usepackage{tikz-cd}
\usepackage{tikz}
\usepackage{cleveref}
\usepackage{booktabs}
\usepackage{multirow}
\usepackage[linesnumbered,ruled]{algorithm2e}
\usepackage{wrapfig}

\usepackage{placeins}
\usepackage{subcaption}
\newtheorem{innercustomthm}{Theorem}
\newenvironment{customthm}[1]
  {\renewcommand\theinnercustomthm{#1}\innercustomthm}
  {\endinnercustomthm}
\newcommand{\bb}{\mathbf}

\DeclareCaptionLabelFormat{andtable}{#1~#2  \&  \tablename~\thetable}

\usepackage[colorinlistoftodos,
	textsize=scriptsize,
	linecolor=red!30,
	bordercolor=red!30,
	backgroundcolor=red!10]{todonotes}

\newtheorem{definition}{Definition}[section]
\newtheorem{prop}{Proposition}[section]
\newtheorem{theorem}{Theorem}[section]
\title{Learning ODEs via Diffeomorphisms for Fast and Robust Integration}
\author{Weiming Zhi$^{1}$ \thanks{Correspondence to \url{weiming.zhi@sydney.edu.au}} \And Tin Lai$^{1}$ \And Lionel Ott$^{2}$ \And Edwin V. Bonilla $^{3}$ \And Fabio Ramos$^{1,4}$\\
$^{1}$ School of Computer Science, the University of Sydney, Australia\\
$^{2}$ Autonomous Systems Lab, ETH Zurich, Switzerland\\
$^{3}$ CSIRO’s Data61, Australia\\
$^{4}$ NVIDIA, USA
}

\begin{document}

\maketitle

\begin{abstract}
Advances in differentiable numerical integrators have enabled the use of gradient descent techniques to learn ordinary differential equations (ODEs). 
In the context of machine learning, differentiable solvers are central for Neural ODEs (NODEs), a class of deep learning models with continuous depth, rather than discrete layers. However, these integrators can be unsatisfactorily slow and inaccurate when learning systems of ODEs from long sequences, or when solutions of the system vary at widely different timescales in each dimension. In this paper we propose an alternative approach to learning ODEs from data: we represent the underlying ODE as a vector field that is \emph{related} to another \emph{base} vector field by a differentiable bijection, modelled by an invertible neural network. By restricting the base ODE to be amenable to integration, we can drastically speed up and improve the robustness of integration. We demonstrate the efficacy of our method in training and evaluating continuous neural networks models, as well as in learning benchmark ODE systems. We observe improvements of up to two orders of magnitude when integrating learned ODEs with GPUs computation.

\end{abstract}

\section{Introduction}
The problem of fitting an ordinary differential equation (ODE) to observed data appears in many disciplines such as biology, chemistry and physics \cite{ODEsBook,pnas}. 
In the context of machine learning and, in particular, neural networks, this problem arises within the framework of 
\emph{Neural ODEs} \cite{chen2018neuralode}, a family of continuous-depth deep learning models that parameterize the dynamics of hidden states using neural networks. Recent developments in learning ODEs generally use differentiable adaptive step-size numerical integrators to learn these dynamics \cite{speed_grad, second_order}. 

\begin{wrapfigure}[16]{R}{0.3\textwidth}
\centering
\begin{subfigure}{.14\textwidth}
  \centering
  \includegraphics[width=\linewidth]{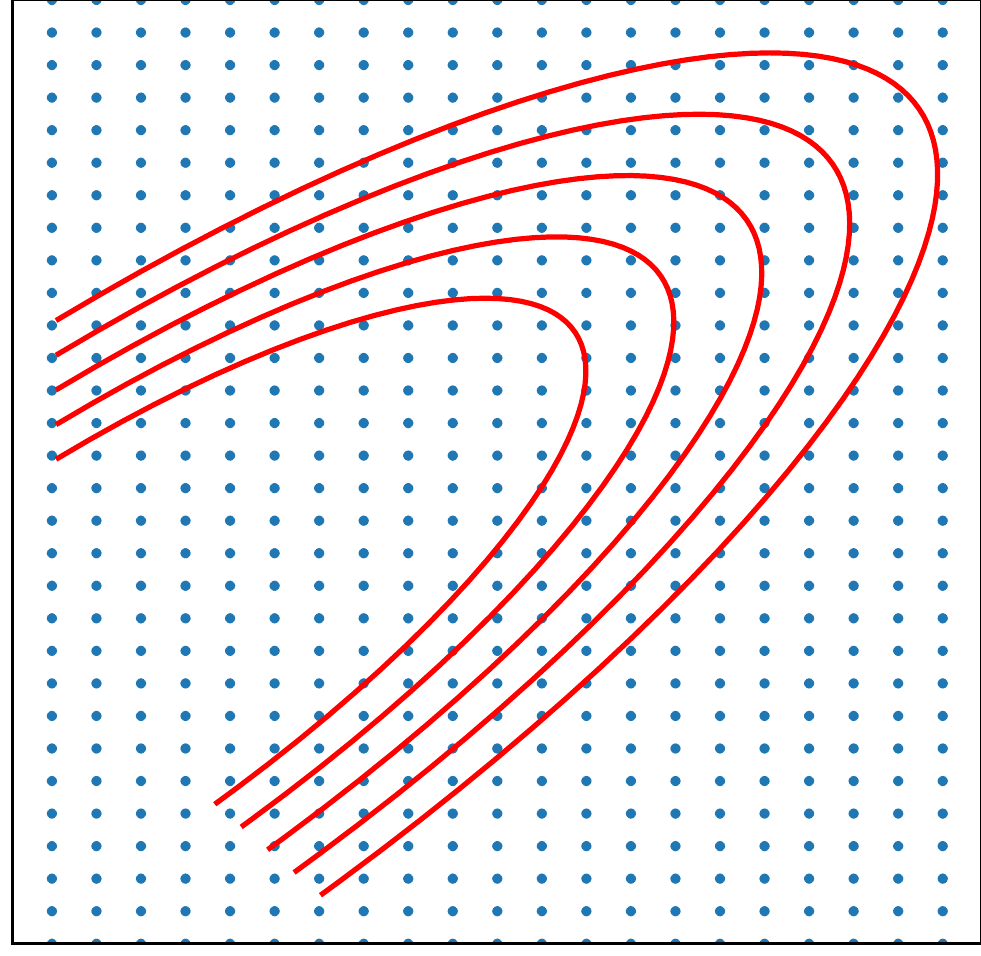}  
\end{subfigure}%
\begin{subfigure}{.14\textwidth}
  \centering
  \includegraphics[width=\linewidth]{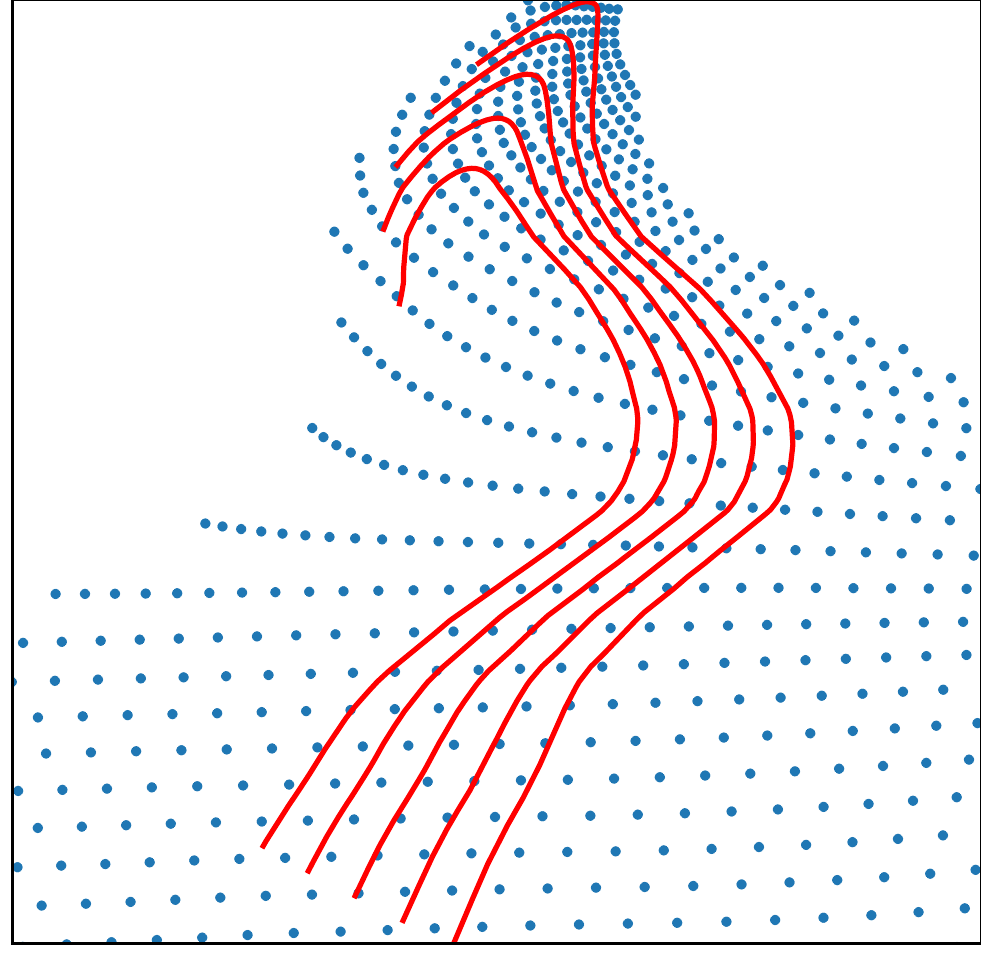}  
\end{subfigure}%
\caption{Related vector fields can be thought of as a vector field that is ``morphed'' into another. (Left) Integral curves (red) of the vector field of a Linear ODE overlaid on grid points (blue); (Right) Morphed integral curves and grid.}
\label{morph}
\end{wrapfigure}

To accurately integrate ODEs with rapidly varying solutions over time, the step sizes taken by the numerical integrator can become exceedingly small. Furthermore, to roll out long sequences of learned ODEs, the neural network dynamics model is queried sequentially at each step. This can be unsatisfactorily slow for time critical applications, such as those in robot control, and is known to suffer from numerical instabilities \cite{Gholami2019ANODEUA,hyperode}, especially if the ODE exhibits stiffness. 
With these 
challenges in mind, we propose an alternative approach to learning ODEs from data. We view the desired target ODE as a vector field that is a ``morphed'' version of an alternative {base vector field} 
via a \emph{diffeomorphism}, i.e., a bijective mapping where both the mapping and its inverse are differentiable. Thus, instead of directly modelling the desired dynamics with a neural network, we use an invertible neural network to learn the diffeomorphism that relates the target ODE to the base ODE. Crucially, the base ODE is much more amenable to integration and, therefore, 
we obtain a solution to this simpler ODE and pass it through the bijection  
to obtain a solution to the (more complex) target ODE. \Cref{morph} shows an example of related vector fields.


We investigate the use of base ODEs that are (1) linear or (2) modelled by a neural network, allowing for integration with accurate numerical integrators at reasonably large step-sizes. 
Restricting the base ODE to be linear allows the use of closed-forms solutions, providing major speed benefits. In this setup, when integrating trajectories of reasonable length, at no degradation of performance, we can achieve a drastic speedup of up to two orders of magnitude when employing GPUs, when compared against existing differentiable integrators with standard settings. We can also restrict the learned ODE to be provably asymptotically stable by adding simple constraints to the linear base ODE. Alternatively, when additional flexibility is required, we remove the restrictive assumption of a linear base ODE and model the dynamics using a neural network. We empirically show that in this setup, we can improve the performance of learning stiff ODEs compared to existing differentiable integrators, even when we use a much smaller neural network for the base ODE. 

In summary, our main contributions are:
\begin{enumerate}
    \item a novel paradigm to learn ODEs from data: invertible neural networks are trained to morph the target ODE to an alternative base ODE that can be more tractably integrated;
    \item analysis of the base as (i) a linear ODE and (ii) a non-linear ODE with learned dynamics and restricted to take large steps. In the linear case, we demonstrate how to obtain closed-form integrals, providing significant speed-ups. In the non-linear case, we demonstrate learning challenging and stiff ODEs, with smaller networks;
    \item a principled method to enforce asymptotic stability of learned ODEs, by adding restrictions to the base ODE. 
    
\end{enumerate}
Proofs and additional details can be found in the supplement.

\section{Related Work}
Our work addresses the problem of learning ODEs that match observations which is typically encountered when learning with a class of continuous-depth neural network model, \emph{neural ODEs} \cite{chen2018neuralode}. Our approach also leverages learnable invertible function approximators in a manner that resembles normalising flows. Here, we briefly outline the related work around learning of ODEs, Neural ODEs, invertible neural networks and normalising flows.

\textbf{Neural ODEs and Learning of ODEs:} 
Neural ODEs are neural network models which model the hidden state as continuous ODEs rather than discrete layers \cite{chen2018neuralode, disect}. Chen et al. proposed to model the dynamics of this information flow by another neural network, which is trained by differentiating integrators via the adjoint method \cite{adjoint} for tractable memory usage. More memory efficient alternatives have also been introduced \cite{mali}. In particular, neural network models which incorporate an ODE, such as latent ODEs \cite{Rubanova2019LatentOF} and neural CDEs \cite{NCDE}, have found application in time-series tasks. Subsequent strategies have been introduced to improve the training of these models, including augmenting the ODE state-space \cite{anode}, regularisation techniques \cite{finlay20a}, and hyper-network extensions \cite{Anodev2, hyperode}. Gaussian processes have also been used in earlier work to model differential equations \cite{simosarkka,Raissi2018NumericalGP}. At the core of neural ODE models is a differentiable integrator which is used to learn the underlying ODE. Our proposed approach improves the learning of the underlying ODE, and is compatible with models that incorporate learnable ODEs. We shall empirically show in \cref{sec:cont_models} our method used to learn dynamics within continuous-depth models. 
The term ``neural ODE'' has typically been used in the literature \cite{Anodev2,hyperode} to refer to neural networks that incorporate ODEs, including the original work in  \cite{chen2018neuralode}. However, ``neural ODE'' has occasionally been used to refer to an ODE with dynamics parameterised by a neural network \cite{norcliffe2021neural}. We follow the former convention. 

\textbf{Invertible neural networks and Normalising Flows:}
Invertible neural networks (INN) are a class of function approximators that learn bijections where the forward and inverse mapping and their Jacobians can be efficiently computed \cite{invertible}. INN are typically constructed by invertible building blocks, such as those introduced in \cite{autoreg,REALNVP,splineflows}. Advances in INNs are largely motivated by normalising flows \cite{normalising, Papamakarios2019NormalizingFF}, an approach to construct a flexible probability distribution by finding a differentiable bijection, or diffeomorphism, between the target distribution and a base distribution. Our approach is similar in spirit to normalising flows, as we analogously aim learn a diffeomorphism that relates the vector fields of the target ODE and some base ODE. However, unlike normalising flows, we do not require the burdensome computation of Jacobian determinants \cite{tract1}. A separate line of work, broadly characterised as \emph{continuous normalising flows}, use ODEs to build invertible approximators \cite{Grathwohl2019FFJORDFC,chen2018neuralode}. Our work proposes the opposite where invertible approximators are used to learn ODEs.



\section{Preliminaries}
We shall introduce learning Ordinary differential equations (ODEs) with neural networks. We briefly present background on invertible neural networks, a core component of our method. We then describe the notions of tangent spaces and pushforwards, which will be used to elaborate our method.  
\subsection{Learning ODEs with neural networks}\label{ode_learn}
ODEs are central in neural ODEs, a class of continuous-depth neural network models. Many problems in science and engineering can also be described by ordinary differential equations of the form:
\begin{align}
    \bb{y}'(t)=f(\bb{y}(t),t), && \bb{y}(0)=\bb{y}_{0},
\end{align}
where $t$ is time, $\bb{y}(t)$ are the states at time $t$, and $f$ provides the dynamics. We use a neural network $f_{\omega}$ with parameters $\omega$ to model the dynamics. We shall henceforth drop the explicit dependence on time, and consider the autonomous ODEs given by $\bb{y}'(t)=f(\bb{y}(t))$. Non-autonomous ODEs, which explicitly depend on time, can be equivalently expressed as autonomous ODEs by adding a dimension to the states $\bb{y}$. Although there have been attempts to explicitly condition the network weights on time \cite{davis2020time}, the dominant approach in the neural ODE literature is simply to append time to the network input. For an initial condition $\bb{y}_{t_{0}}$ at start $t_{0}$, and some end time $t_{e}$, a solution of the ODE can be evaluated by a numerical integrator (ODESolve), such as Euler's or Runge-Kutta methods \cite{rk_num}:
\begin{align}
    \bb{y}(t_{e})=\bb{y}_{t_{0}}+\int^{t_{e}}_{t_{0}}f_{\bb{\omega}}(\bb{y}(t))\mathrm{d}t=\mathrm{ODESolve}(f_{\bb{\omega}},\bb{y}_{t_{0}},t_{e}).
\end{align}
The learning problem involves estimating, with $f_{\omega}$, the dynamics of the ODE, provided $n$ observations $\bb{y}^{obs}_{t_{1}}\ldots\bb{y}^{obs}_{t_{n}}$ at specified times. We can learn the ODE by optimising the parameters $\omega$ to minimise a loss between the observations at the given times and the integrated ODE, 
 $\ell(\bb{\omega})=\mathrm{Loss}(\bb{y}^{obs}_{t_{i}}, \bb{y}(t_{i}))$ for $i=1,\ldots,n$. Advances in the neural ODE literature have introduced differentiable numerical integrators, which allow gradient descent optimisation to be applied. By using the adjoint sensitivity method as outlined in \cite{chen2018neuralode}, the gradients of adaptive integrators can be obtained in a memory tractable manner, without differentiating through the operations of the integrator.

\subsection{Invertible Neural Networks}
A key building block of our method is the invertible neural network (INN). Invertible neural networks are a class of function approximators which learn differentiable bijections. INNs can be trained on a forward mapping, and get the inverse mapping with no further labour, owing to the definition of their architecture. Throughout this paper, we use INNs of the type described in \cite{REALNVP}. 
The basic unit is a reversible block, where inputs are split into two halves, $\bb{u}_{1}$ and $\bb{u}_{2}$, and the outputs $\bb{v}_{1}$ and $\bb{v}_{2}$ are:
\begin{align}
    \bb{v}_{1}=\bb{u}_{1}\odot \exp(s_{2}(\bb{u}_{2}))+t_{2}(\bb{u}_{2}), && 
    \bb{v}_{2}=\bb{u}_{2}\odot \exp(s_{1}(\bb{u}_{1}))+t_{1}(\bb{u}_{1}),
\end{align}
where $\odot$ indicates element-wise multiplication, and $t_{1}$, $t_{2}$ and $s_{1}$, $s_{2}$ are functions modelled by fully-connected neural networks with non-linear activations. These expressions are clearly invertible:
\begin{align}
    \bb{u}_{1}=(\bb{v}_{1}-t_{2}(\bb{u}_{2}))\odot\exp(-s_{2}(\bb{u}_{2})), && \bb{u}_{2}=(\bb{v}_{2}-t_{1}(\bb{u}_{1}))\odot\exp(-s_{1}(\bb{u}_{1})).
\end{align}
The functions $t_{1}$, $t_{2}$ and $s_{1}$, $s_{2}$ themselves are not required to be invertible, and can be modelled by regular neural networks. Although further developments have given rise to methods of constructing better performing invertible blocks, such as those in \cite{splineflows}, the coupling-based block introduced here is fast, sufficiently flexible, and has been shown to be a universal diffeomorphism approximator \cite{universal}.   

\subsection{Tangent Spaces and Pushforwards}

Here we briefly introduce the differential geometry notions of tangent spaces and pushforwards, which will be later used to elaborate on geometric concepts of our method.  

\textbf{Tangent Spaces:} A manifold is a space that locally resembles Euclidean space. Throughout this paper, all manifolds will be assumed to be differentiable, with defined \emph{tangent spaces}. For an $n$-dimensional manifold $\mathcal{M}$, at a point $\bb{p}\in\mathcal{M}$, the tangent space $T_{\bb{p}}\mathcal{M}$ is an $n$-dimensional real vector space, where each element passes $\bb{p}$ tangentially and is referred to as a \emph{tangent vector}. The tangent space provides a higher-dimensional analogue of a tangent plane at a point on a surface. The collection of tangent spaces for all points on $\mathcal{M}$ is known as the \emph{tangent bundle} denoted by $T\mathcal{M}$.

\textbf{Pushfoward:} For a mapping $F:\mathcal{M}\rightarrow\mathcal{N}$ between two manifolds, $\mathcal{M}$ and $\mathcal{N}$, the \emph{pushforward} by $F$ is a linear mapping between the tangent spaces of the manifolds, $D_{\bb{p}}F: T_{\bb{p}}\mathcal{M}\rightarrow T_{F(\bb{p})}\mathcal{N}$. Tangent vectors at $\bb{p}$ in the domain $\mathcal{M}$ can be mapped to tangent vectors at the corresponding point $F(\bb{p})$ in the codomain $\mathcal{N}$ via the pushforward. This can be computed by the matrix product of the Jacobian of $F$ at $\bb{p}$ and a tangent vector at $\bb{p}$.  

\section{Methodology}
To elaborate on our method, we shall study ODEs as vector fields. The dynamics of an ODE, $f$, can be viewed as a vector field, and solutions as integral curves of the vector field \cite{LeeManifolds}. First, we introduce the learning of vector fields which are \emph{related} to another \emph{base} vector field. Second, we describe possible choices for such \emph{base} vector fields.

\begin{figure}[t]
\centering

\begin{subfigure}[b]{0.45\textwidth}
                \centering
                \includegraphics[width=\textwidth]{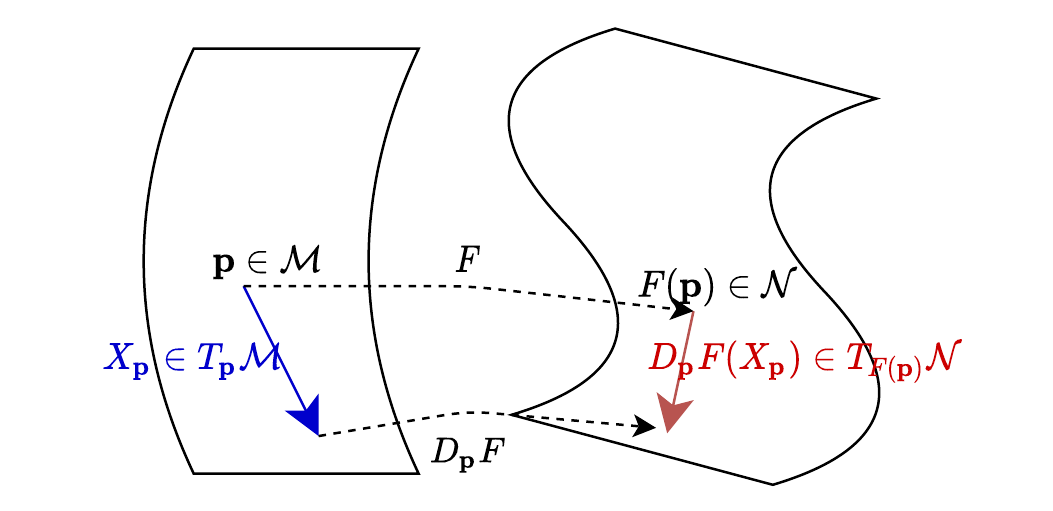}
                \caption{If $F$ maps points $\bb{p}\in\mathcal{M}$ to $F(\bb{p})\in\mathcal{N}$, a single tangent vector at $\bb{p}$, $X_{\bb{p}}\in T_{\bb{p}}\mathcal{M}$, can be mapped to $T_{\bb{F(\bb{p})}}\mathcal{N}$. However, an entire vector field $X$ on $\mathcal{M}$ cannot in general be mapped to a valid vector field on $\mathcal{N}$. The pushforward by a diffeomorphism is a special case where a valid vector field can be obtained.} \label{diag2}
        \end{subfigure}%
\hspace{1em}
\begin{subfigure}[b]{0.45\textwidth}
\begin{tikzpicture}
\centering
  \matrix (m) [matrix of math nodes,row sep=4em,column sep=8em,minimum width=4em,font=\large]
  {
     T\mathcal{M} & T\mathcal{N} \\
     \mathcal{M} & \mathcal{N} \\};
  \path[-stealth]
    (m-1-1) edge node [below] {\large $DF$} (m-1-2)
    (m-2-1) edge node [left] {\large $X$} (m-1-1)       
    (m-2-1.east|-m-2-2) edge node [above] {\large $F$} (m-2-2)
    (m-2-2.south west) edge node [below] {\large $F^{-1}$} (m-2-1.south east)
    (m-2-2) edge node [right] {\large$Y$} (m-1-2);
\end{tikzpicture}\caption{If vector fields $X$ and $Y$ on manifolds $\mathcal{M}$ and $\mathcal{N}$ respectively are \emph{related} by diffeomorphism $F$, then they are related via the pushforward of $F$. If $Y$ is unknown, we have another path to evaluate $Y$ by  $\mathcal{N}\xrightarrow[]{F^{-1}}\mathcal{M}\xrightarrow[]{X}T\mathcal{M}\xrightarrow{DF}T\mathcal{N}$.} \label{diag1}
\end{subfigure}%
\caption{Vector fields related by a diffeomorphism $F: \mathcal{M}\rightarrow \mathcal{N}$ allow the pushforward of $F$ to act on vector fields on $\mathcal{M}$ producing vector fields on $\mathcal{N}$.}
\end{figure}

\subsection{Related Vector Fields for ODE Learning}
A vector field $X$ defined on manifold $\mathcal{M}$ is a function that assigns a tangent vector $X_{\bb{p}}\in T_{\bb{p}}\mathcal{M}$ to each point $\bb{p}\in\mathcal{M}$. Intuitively, our aim is to construct a mapping $F$ which shapes the manifold where a \emph{base} vector field $X$ is defined, such that the pushforward of $X$ by $F$ extrinsically appears ``morphed'' to match the data. A question that arises is: \textbf{\emph{what are the requirements of these mappings, for the ``pushed forward'' vector field to be valid?}}


Provided a mapping between manifolds $F: \mathcal{M}\rightarrow\mathcal{N}$, we can push a single vector, $X_{\bb{p}}\in T_{\bb{p}}\mathcal{M}$, to the tangent space of $\mathcal{N}$ at $F(\bb{p})$, $T_{F(\bb{p})}\mathcal{N}$, via the \emph{pushforward}, $D_{\bb{p}}F(X_{\bb{p}})$. \Cref{diag2} sketches out an example of a ``pushing forward'' a tangent vector between tangent spaces. However, this notion does not extend in general to vector fields. If mapping $F$ is injective and non-surjective, the pushforward of $X$ outside the image of $F$ is not defined. On the other hand, if mapping $F$ is surjective and non-injective, there may be multiple differing pushforwards given for a point. In special cases when the pushforward of mapping $F$ defines a valid vector field on the codomain $\mathcal{N}$, the vector field and its pushforward are known to be $F$-related.

\begin{definition}[Related vector fields]
Let $F:\mathcal{M}\rightarrow\mathcal{N}$ be a smooth mapping of manifolds. A vector field $X$ on $\mathcal{M}$ and a vector field $Y$ on $\mathcal{N}$ are related by $F$, or \emph{$F$-related}, if for all $\bb{p}\in\mathcal{M}$,
\begin{equation}
    D_{\bb{p}}F(X_{\bb{p}})=Y_{F(\bb{p})}
\end{equation}
\end{definition}

Related vector fields arise in particular when $F$ is a bijective mapping, where both the mapping itself and its inverse are differentiable, i.e. a \emph{diffeomorphism}. 

\begin{prop} [Proposition 8.19 in \cite{LeeManifolds}]
Suppose $F:\mathcal{M}\rightarrow\mathcal{N}$ is a diffeomorphism between smooth manifolds $\mathcal{M}$, $\mathcal{N}$. For every vector field $X$ on $\mathcal{M}$, there is a unique vector field $Y$ on $\mathcal{N}$ that is $F$-related to $X$.
\end{prop}

By considering the $F$-related properties of vector fields, we have a pathway to define unknown vector fields using the pushforward of $F$, as shown in \cref{diag1}. If vector field $X$ on $\mathcal{M}$ is $F$-related to some vector field $Y$ on $\mathcal{N}$, instead of directly evaluating the vector field $Y$, we can instead obtain tangent values for any $\bb{q}\in\mathcal{N}$, via $\mathcal{N}\xrightarrow[]{F^{-1}}\mathcal{M}\xrightarrow[]{X}T\mathcal{M}\xrightarrow{DF}T\mathcal{N}$. Therefore, the vector attached by vector field $Y$ is, 
\begin{equation}
Y_{\bb{q}}=D_{F^{-1}(\bb{q})}F(X_{F^{-1}(\bb{q})})=J_{F}(F^{-1}(\bb{q}))X_{F^{-1}(\bb{q})} \text{, for each }\bb{q}\in\mathcal{N}, \label{pushforwardeq}
\end{equation}
where $J_{F}$ is the Jacobian of diffeomorphism $F$. One would naturally ask: \textbf{\emph{why would it be beneficial to construct a desired vector field $Y$ in the form of \cref{pushforwardeq}?}} 

We shall answer this by considering integral curves on $Y$, which represent solutions to the ODE associated with $Y$. An integral curve of $Y$ on $\mathcal{N}$ is a differentiable curve $\bb{y}:\mathbb{R}\rightarrow\mathcal{N}$, whose velocity at each point is equal to the value of $Y$ at that point, i.e. $\bb{y}'(t)=Y_{\bb{y}(t)}\in T_{\bb{y}(t)}\mathcal{M}$, for all $t\in\mathbb{R}$. The integral curves of $F$-related vector fields are also linked by $F$: integral curves on one vector field are mapped to the other via a single pass through $F$, and evaluation of Jacobian $J_{F}$ is not required.

\begin{prop} [Proposition 9.6 in \cite{LeeManifolds}]
Suppose $X$ and $Y$ are vector fields on manifolds $\mathcal{M}$ and $\mathcal{N}$ respectively. $X$ and $Y$ are related by mapping $F:\mathcal{M}\rightarrow\mathcal{N}$ if and only if for each integral curve $\bb{x}:\mathbb{R}\rightarrow\mathcal{M}$, $\bb{y}=F(\bb{x})$ is an integral curve of $Y$.
\end{prop}

In the ODE learning problem outlined in \cref{ode_learn}, during both training and inference, we need to obtain an integral curve $\bb{y}$ of the vector field $Y$ either by numerical integration on $Y$, or by $\bb{y}=F(\bb{x})$, where $\bb{x}$ denotes the corresponding integral curve of $X$, related to $Y$ via the diffeomophism $F$. The Jacobian of $F$ does not need to be evaluated when we are working with the integral curves.  

If integral curves of $X$ can be found in a more efficient, or less error-prone manner, than by numerically integrating curves of $Y$, we can leverage the relationship $\bb{y}=F(\bb{x})$ for ODE learning. This can be done by an invertible neural network, with parameters $\theta$, to parameterise a diffeomorphism, $F_{\theta}$, which maps between $\mathbb{R}^{n}$ and some $n$-dimensional manifold that is diffeomorphic to Euclidean space. We denote the base ODE associated with the $F$-related vector field of our target as $\bb{x}'(t)=g_{\varphi}(\bb{x}(t))$, where $\varphi$ parameterises the base ODE. We can then use the target ODE within some learning problem, such as minimising the mean absolute error (MAE) between the target ODE and observations:
\begin{equation}
    \ell(\theta,\varphi)=MAE \Big(\bb{y}_{t_{i}}^{obs},F_{\theta}(F_{\theta}^{-1}(\bb{y}_{0})+\int^{t_{i}}_{0}g_{\varphi}(\bb{x}(t))\mathrm{d}t)\Big),
\end{equation}
where $\bb{y}_{0}$ is an initial condition for the system, and $\bb{y}_{t_{i}}^{obs}$ are observed data points at times $t_{i}$. By \cref{pushforwardeq}, the learned ODE can then be written as $\bb{y}'(t)=J_{F_{\theta}}(F_{\theta}^{-1}(\bb{y}(t)))g_{\varphi}(F_{\theta}^{-1}(\bb{y}(t)))$. Provided an initial solution $\bb{y}_{0}$, we can solve via $\bb{y}(t)=F_{\theta}(F_{\theta}^{-1}(\bb{y}_{0})+\int^{t}_{0}g_{\varphi}(\bb{x}(t))\mathrm{d}t)$. In practice, we are often required to evaluate an entire trajectory, ie. $\bb{y}(t)$ at multiple times $t_{1},\ldots,t_{end}$ with one initial $\bb{y}_{0}$, as outlined in Algorithm \ref{integration}. This allows us to batch up the pass through $F_{\theta}$, which makes this highly efficient when executed on a GPU. The benefits of our method are apparent when it is advantageous to integrate the base ODE and then pass the solution through the diffeomorphism, $F_{\theta}$, rather than numerically integrate the target ODE. Next, we investigate two choices of base ODE: (1) Linear ODE; (2) Non-linear ODE amenable to numerical integration.
\begin{algorithm}[t]
    \SetKwInOut{Input}{Input}
    \SetKwInOut{Output}{Output}
    \Input{$F_{\theta}$, $g_{\varphi}$, $\bb{y}_{0}$, $t_{1},\ldots,t_{end}$}
    \Output{$\bb{y}(t_{1}),\ldots,\bb{y}(t_{end})$ }
    $\bb{x}_{0}\gets F^{-1}_{\theta}(\bb{y}_{0})$\\
    $\bb{x}(t_{i})\gets \bb{x}_{0}+\int^{t_{i}}_{0}g_{\varphi}(\bb{x}(t))\mathrm{d}t$, for $i=1,\ldots,end$ \tcp*{The integral either admits a closed-form solution or is easily numerically integrated.}
    \caption{Efficient integration of learned ODEs}\label{integration}
    $\bb{y}(t_{1}),\ldots,\bb{y}(t_{end})\gets F_{\theta}(\bb{x}(t_{1}),\ldots,\bb{x}(t_{end}))$ \tcp*{The pass through the INN can be batched, and efficiently computed on GPUs.}

\end{algorithm}


\subsection{Linear ODE as Base: Fast Integration and Easy Enforcement of Asymptotic Stability}
We can speed-up integration significantly by modelling the base as a Linear ODE. Linear ODEs are in the form of $\bb{x}'(t)=A\bb{x}(t)$, where $\bb{x}(t)\in\mathbb{R}^{n}$ are $n$-dimensional variables, and $A\in\mathbb{R}^{n \times n}$. Linear ODEs can be solved very efficiently as they admit closed-form solutions. Provided an initial solution $\bb{x}_{0}$, the solution of $\bb{x}(t)$ and the target ODE $\bb{y}(t)$ are then
\begin{align}
\bb{x}(t)=\sum_{k=1}^{n}(\bb{l}_{k}\cdot \bb{x}_{0})\bb{r}_{k}\exp(\lambda_{k}t), && \bb{y}(t)=F_{\theta}(\bb{x}(t)),
\end{align}
where $\bb{l}_{k}$, $\bb{r}_{k}$ and $\lambda_{k}$ are the corresponding left, right eigenvectors and eigenvalue of matrix $A$ respectively. We learn the eigenvalues and eigenvectors or matrix $A$ jointly with diffeomorphism $F_{\theta}$. 

Linear ODEs are also interesting because their long-term behaviour, which is determined by their eigenvalues, is easy to analyse.
We shall see how this property allows us to craft the long-term behaviour of the desired target ODE. 
In particular, in many applications, consideration is given to the asymptotic properties of ODEs, namely what happens to the solutions after a long period of time. Will the solution converge to equilibrium points, periodic orbits, or diverge and fly off? Non-linear ODEs with neural network dynamics typically do not restrict the learned ODE to be provably stable. Learning ODEs with our method provides a straightforward way to restrict the ODE to be asymptotically stable. We begin by defining equilibrium points and asymptotic stability of first order ODEs.
\begin{definition}[Equilibrium point]
An equilibrium point $\bb{y}^{*}$ of an ODE $\bb{y}'(t)=f(\bb{y}(t))$, is a point where $f(\bb{y}^{*})=0$. 
\end{definition}
\begin{definition}[Asymptotic stability]
An ODE $\bb{y}'(t)=f(\bb{y}(t),t)$ is asymptotically stable if for every solution $\bb{y}(t)$, there exists a $\delta>0$, such that whenever $\lvert\lvert \bb{y}(t_{0})-\bb{y}^{*} \lvert\lvert< \delta$, then $\bb{y}(t)\rightarrow \bb{y}^{*}$ as $t\rightarrow \infty$, where $\bb{y}^{*}$ is some equilibrium point.  
\end{definition}
Intuitively, asymptotically stable systems of ODEs will always settle at some equilibrium points after a long period of time. In the context of vector fields related by a diffeomorphism, the asymptotic stability properties of the ODEs are shared. 
\begin{theorem}
Suppose two ODEs $\bb{x}'(t)=g(\bb{x}(t))$, $\bb{y}'(t)=f(\bb{y}(t))$ are related via $\bb{y}(t)=F(\bb{x}(t))$, where $F$ is a diffeomorphism. If the former ODE is asymptotically stable with $n_{e}$ equilibrium points $\bb{x}^{*}_{1},\ldots,\bb{x}^{*}_{n_{e}}$, then the latter ODE is also asymptotically stable, with equilibrium points $F(\bb{x}^{*}_{1}),\ldots,F(\bb{x}^{*}_{n_{e}})$.
\end{theorem}
Therefore, if we can restrict the base ODE to be asymptotically stable, then the target ODE learned by our method is also asymptotically stable. When the base is an $n$ dimensional linear ODE, we can restrict it to be asymptotically stable by directly learning the eigenvalues, $\lambda_{i}$ for $i=1,\ldots,n$, and constraining them to be negative, i.e. $\lambda_{i}<0$ for $i=1,\dots,n$. This can be done by $\lambda_{i}=-(s_{\lambda_{i}})^{2}-\varepsilon I$, where $\varepsilon$ is a small positive constant, and learning $s_{\lambda_{i}}$ instead of learning the eigenvalues.
\subsection{Non-linear ODE as Base: Improved Robustness for `Difficult' ODEs}

Using linear systems as base ODEs provides a dramatic increase in speed at the cost of flexibility. We observe that the computation overhead of a single backward pass $F^{-1}_{\theta}$ and a batched single forward pass $F_{\theta}$ is minimal when compared with numerical integration. When the ODE is difficult to learn and only a moderate speed-up is required, we can also parameterise the dynamics of the base ODE using a neural network. This is particularly appealing for ODEs which are considered stiff. 

Although there is no precise definition of stiffness, a common characteristic of stiff ODEs is a rapid varying of the solution in time at different orders of magnitude across state dimensions.
Existing differentiable explicit integrators are unable to learn these ODEs. Learning the target ODE will require step-sizes that are exceedingly small at even large tolerances. Adaptive step-size solvers will encounter arithmetic underflow, while fixed step-size solvers are struggle to accurately integrate \cite{stiff_odes}. We restrict the base ODE to be solved with an Euler method integrator, and learn it jointly with the diffeomorphism. The burden of accurately representing the stiff dynamics is shared by $F_{\theta}$, providing added flexibility. The diffeomorphism is observed to learn to relate the target ODE to an ODE that is amenable to integration, where the timescales across dimensions do not differ greatly at the same time. This set-up allows us to learn stiff systems to an accuracy that cannot be achieved by  typical integrators, even when the neural network model of the base ODE is much smaller than the neural network used to directly learn the target ODE, providing a speed-up during integration.

\section{Experimental Results}
We empirically evaluate the ability of our method to speed up the integration of learned ODEs, along with the robustness of integration when learning potentially stiff ODEs. Throughout this section, we compare the error and integration times of our method against a variety of solvers. Including fixed step-size solvers: Euler's, midpoint, and Runge-Kutta 45 (RK4), and the adaptive step-size solvers Dormand–Prince 5 (DOPRI5) and Dormand–Prince 8 (DOPRI8). For all fixed step-size solvers, we set the step-size equal to the smallest time increment for which we require outputs. We augment the ODE states in accordance to \cite{anode} in all of the ODEs trained during the experiments, except when recreating results from \cite{Rubanova2019LatentOF} in \cref{sec:cont_models}, where we use the implementation from the original authors. For all adaptive step-size solvers, we set absolute and relative tolerances to $10^{-5}$. The differentiable solvers are implemented in the \emph{torchdiffeq} library, with neural networks implemented with Pytorch \cite{torch}. Additional details on experimental setup are available in the supplement.

\subsection{Substantial Integration Speed-up by Learning with a Linear ODE Base}
We test our hypothesis that the availability of a closed-form expression for the integral, when using a linear base ODE, provides a substantial speed-up of integration. We evaluate on learning synthetic ODE systems, real-world robot demonstrations, and within a \emph{Latent ODE} \cite{Rubanova2019LatentOF}. Here, we report performance and integration times. Training times can be found in the supplement. 


\begin{wrapfigure}[14]{R}{0.34\textwidth}
\centering
\begin{subfigure}{.165\textwidth}
  \centering
  \includegraphics[width=\linewidth]{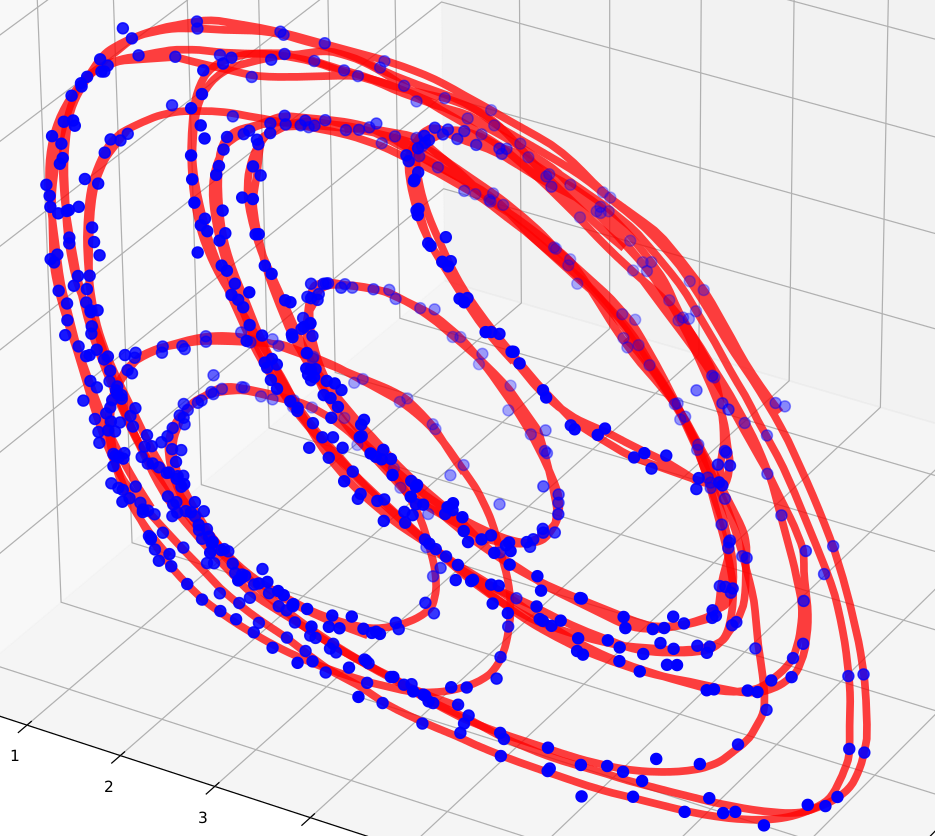}  
\end{subfigure}%
\begin{subfigure}{.165\textwidth}
  \centering
  \includegraphics[width=\linewidth]{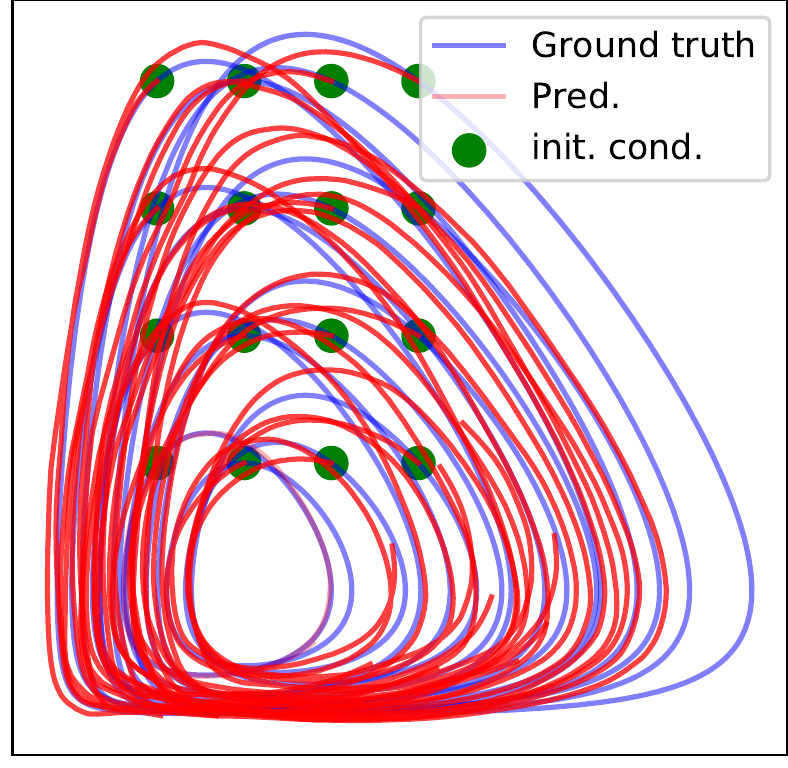}  
\end{subfigure}%
\caption{Learning the 3D Lotka–Volterra. (Left) Interpolating (red) data (blue); (Right) Generating trajectories (red) at unseen initial conditions and the ground truth (blue).}\label{LV}
\end{wrapfigure}

\subsubsection{Learning 3D Lotka-Volterra}
We train and evaluate models on data from the 3D Lokta-Volterra system, which models the dynamics of predator-prey populations. The data is corrupted by white noise with standard deviation of $0.05$. We train our model using a linear base ODE, and assess the capability of our model in interpolating the data points at $10$x the data resolution, and generalising to $16$ unseen initial conditions to integrate trajectories, also at $10$x the data resolution. The integrated trajectories of the same time duration for interpolation and generalisation to novel initial conditions have similar integration run-times. Hence, we only report the integration time for generalisation. 
\Cref{LV} shows interpolation results and newly generated trajectories, where we see that our model is able  to capture the dynamics of the system. 
%
Furthermore, \cref{tab:Linres} provides 
a quantitative evaluation,  where we see that our method not only achieves better performance in terms of MSE but also is significantly faster than the competing numerical integrators with speed-ups of more than two orders of magnitude.

\subsubsection{A Time Critical Application: Motion Generation After Learning from Demonstrations}
The ability to quickly roll-out trajectories is crucial in robotics settings. In particular, we consider the application of generating new motion trajectories from provided demonstrations. We evaluate the ability of our method in learning end-effector trajectories of robot manipulators. The goal is to learn an ODE system where trajectories integrated at different starting points mimic the shown demonstrations. We use three sets of data from \cite{KhansariZadeh2011LearningSN}: (i) 7 demonstrations of drawing ``S'' shapes on a flat surface; (ii) 14 demonstrations of placing a cube on a shelf; (iii) 12 demonstrations of drawing out large ``C''-shapes. We use $70\%$ of the data for training, and test our generalisation capabilities on the remaining demonstrations. In these datasets, the motion converges to equilibrium points. Hence, we constrain the learned ODE to be asymptotically stable. We report the performance and run-times of generalising to new starting points in \cref{tab:Linres}. We see that our approach is competitive in the quality of generalised trajectories, while achieving speed-ups  of more than  two-orders of magnitude.


\subsubsection{ODE Learning for Continuous Deep Learning Models}\label{sec:cont_models}

We evaluate our method as a component of \emph{Latent ODEs} \cite{Rubanova2019LatentOF}, a continuous-depth deep learning model. Latent ODEs embeds the time series observations as hidden states via an encoder-decoder. An ODE, with dynamics parameterised by a neural network, is fit on the hidden states, using a differentiable integrator. By assuming the hidden states follow some continuous time dynamics, the model is capable of handling irregularly sampled series. This avoids the need to group observations into equally-timed bins. In our experiments, we apply our method, with a linear ODE base, to learn the dynamics governing the hidden states. We report results for reconstructing the periodic curves and the human activity classification problem, which were used in the original latent ODE paper \cite{Rubanova2019LatentOF}, as well as an additional ECG classification problem. We reconstruct the periodic curves for 100 and 1000 time-steps. The performance and times spent on integrating the hidden state dynamics are reported in \cref{latent_res}. In our comparisons, the latent ODE is set up according to the original paper. We see that by leveraging the closed-form expression of integrals of the ODEs, we achieve integration times that are hundreds of times faster. We also note that the main computation cost of the integral in our method is the pass through the invertible neural network. The parallel computation capabilities of GPUs allow us to batch the pass at constant cost, whereas the sequential nature of numerical integrators result in a linear increase in run-time, as demonstrated by the differences in integration times of ``Periodic 100'' and ``Periodic 1000''.        




\begin{table}[t]
\resizebox{\textwidth}{!}{%
\begin{tabular}{llllllllll}
\toprule
                    & \multicolumn{3}{c}{3D Lotka-Volterra} & \multicolumn{2}{c}{Imitation S}      & \multicolumn{2}{c}{Imitation cube pick}  & \multicolumn{2}{c}{Imitation C}   \\ \midrule
                   & MSE (I)  & MSE (G) & Time (ms) & MSE (G) & Time (ms)  & MSE (G) & Time (ms)  & MSE (G) & Time (ms) \\\midrule
Ours (Lin)            &   0.143± 0.096       &     1.482± 0.11      &       9.32± 0.37    &    6.11±1.21                     &           6.62± 0.15             &     18.61± 6.21 & 7.12 ± 1.58       & 8.12± 1.58 & 7.53± 0.76

          \\
Euler                       &       4.460± 0.264        &    4.621± 0.123       &     385.55± 14.42      &        10.30±2.90                 &        724.65± 8.32                 &    14.92± 1.38 & 728.40± 9.46        & 7.26± 1.99 &  753.94± 1.44
           \\
Midpoint                      &      0.382± 0.047         &      5.514± 0.120     &     670.37± 31.25      &     10.93±3.25                    &         581.59± 13.27                &    12.89± 1.25  & 1267.16± 13.58      & 6.87± 2.23 & 1305.44± 14.74
           \\
RK4                          &       0.354± 0.005        &      5.630± 0.149     &     1316.11± 30.84      &       10.27±2.99                  &         2501.67± 18.91               &      15.89± 0.89 & 2522.74± 23.14      & 7.59± 2.64 &  1292.27± 21.95
          \\
DOPRI5                        &      0.927± 0.050         &      5.186± 0.365     &      264.67± 17.02     &       10.83±2.76                  &           1277.71± 14.24             &    14.88± 0.86 & 504.04± 12.33     & 7.13± 1.92 &  623.38± 15.57
      
\\ \bottomrule
\end{tabular}%
}\caption{
The mean squared error for interpolation, MSE (I), and generalisation, MSE (G), and mean execution times (± 1 standard deviations) on the 3D  Lotka-Volterra system and the time critical application of motion trajectory generation from demonstrations for our method using a linear base ODE and competing numerical integrators.  
}\label{tab:Linres}%
\end{table}

\begin{table}[t]
\centering
\resizebox{\textwidth}{!}{%
\begin{tabular}{lcccccccc}
\hline
\multicolumn{1}{c}{} & \multicolumn{2}{c}{Periodic 100} & \multicolumn{2}{c}{Periodic 1000} & \multicolumn{2}{c}{Human Activity} & \multicolumn{2}{c}{ECG} \\ \cline{2-9} 
\multicolumn{1}{c}{} & MSE & Int. time (ms) & MSE & Int. time (ms) & Acc. & Int. time (ms) & Acc. & Int. time (ms) \\ \hline
Ours (Lin) & 0.030 & 2.7±  0.6 & \multicolumn{1}{l}{0.008} & 2.8±  0.8 & 0.864 & 4.2±  1.8 & 0.966 & 7.6± 2.5 \\
Euler & 0.040 & 33.7±  2.6 & 0.043 & 326.6±  9.5 & 0.815 & 67.9±  2.9 & 0.963 & 100.0±  2.8 \\
Midpoint & 0.032 & 54.5±  1.8 & 0.074 & 510.1± 15.5 & 0.865 & 114.2±  2.4 & 0.963 & 169.7±  3.8 \\
RK4 & 0.039 & 95.6±  1.6 & 0.052 & 1020.0±  60.0 & 0.857 & 221.2±  4.2 & 0.963 & 325.5±  4.8 \\
DOPRI5 & 0.045 & 83.4±  2.2 & 0.050 & 264.7±  4.6 & 0.869 & 67.9± 5.0 & 0.963 & 123.3±  2.6 \\
DOPRI8 & 0.041 & 99.6±  2.3 & 0.049 & 282.7±  6.4 & 0.724 & 94.8±  1.6 & 0.963 & 171.6±  3.6 \\ \hline
\end{tabular}}
\caption{
The mean squared error and mean integration times (± 1 standard deviations) when using latent ODEs on the tasks of periodic curve reconstruction using 100 and 1000 time-steps and the classification problems of human activity and ECG for our method using a linear base ODE and competing numerical integrators.
}\label{latent_res}
\end{table}

\begin{figure}[]
    \centering
    \includegraphics[width=0.9\textwidth]{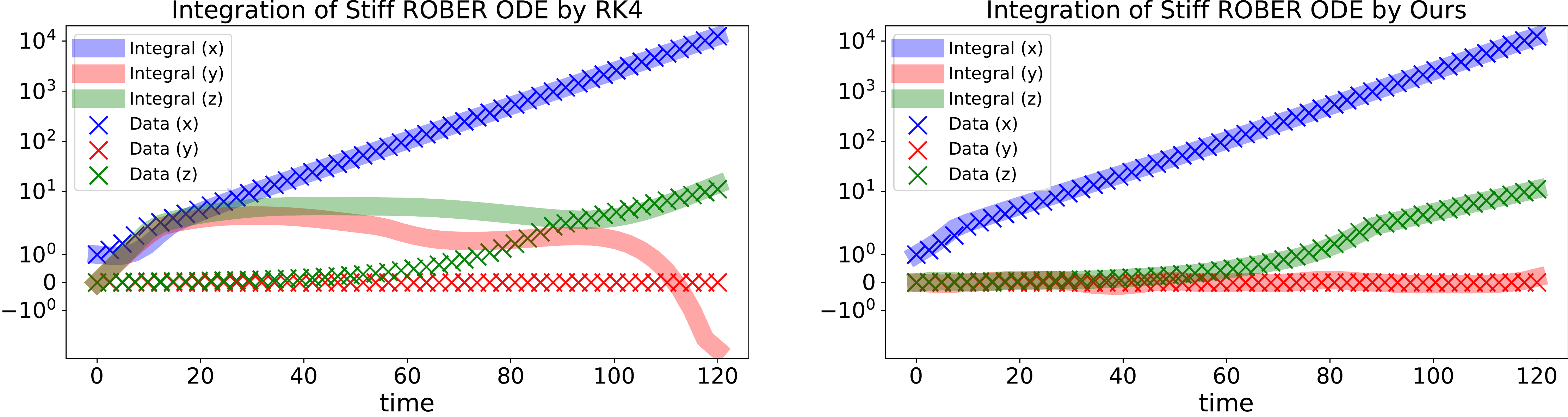}
    \caption{An example of integrating stiff ROBER (log-scale). (Left) When learning stiff ROBER with RK4, we see that integrated curves struggle to match the ground truth at dimensions with small scales. (Right) Learning and integration by our method matches the ground truth better. }
   \label{ROBER_comp}
\end{figure}

\subsection{Robust Integration by Learning with a Non-Linear Neural Network Base}

\begin{figure}
\centering
\begin{subfigure}{.3\textwidth}
  \centering
  \includegraphics[width=0.85\linewidth]{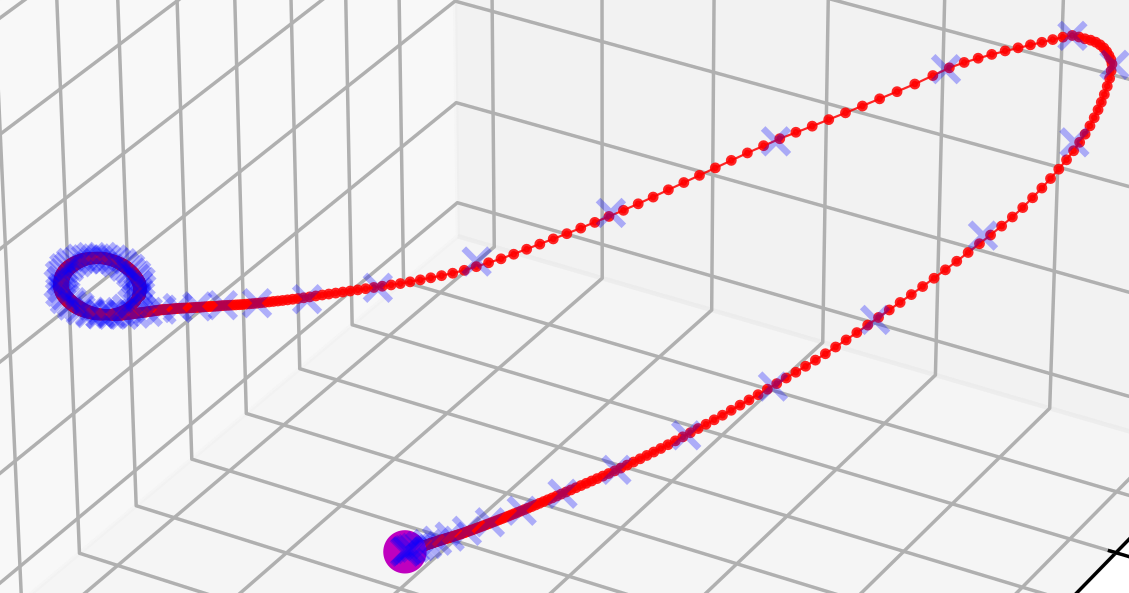}  
\caption{Lorenz system. Data (blue) with generated trajectory (red).}\label{Lorenz}
\end{subfigure}%
\hspace{2em}
\resizebox{0.55\textwidth}{!}{
\begin{tabular}{lllll}
\toprule
               & \multicolumn{2}{l}{Lorenz} & \multicolumn{2}{l}{ROBER} \\
               \midrule
               & MAE          & Time(ms)       & MAE         & Time(ms)        \\
Ours (Non-Lin) &  0.131     &          403.62± 64.89         & 1.33      &  24.60± 1.83 \\
Euler          &  10.99            &   455.67± 13.11         & 2.03      &   26.35± 0.57 \\
Midpoint       &  6.60            &    805.05± 42.95         & 2.49      &    48.89± 3.52 \\
RK4            &  6.81            &    1760.84± 205.54       & 1.60      &  98.68± 4.65 \\
DOPRI5         &  7.55            &    632.22± 83.33         & N/A        &  N/A \\
\bottomrule
\end{tabular}}
\caption{
Results on chaotic (Lorenz) and stiff (ROBER) systems. \textit{Left:} a generated trajectory by our method along with the true underlying data. \textit{Right:} Quantitative evaluation of our method using a non-linear base ODE and competing numerical integrators.
\label{Results_hard}
}
\end{figure}

We test our hypothesis that using a neural network base ODE allows us to learn ODEs that are difficult to integrate or stiff. 
We learn and evaluate models trained on the chaotic Lorenz system, and the stiff Robertson's system \cite{rober} (ROBER). A simple differentiable Euler integrator, with step-size equal to the data time-step size, is used to learn our base ODE. \Cref{Lorenz} illustrates a generated trajectory from the Lorenz system at $10$x the data resolution, which closely resembles the data points (blue). For the stiff ROBER system, attempts to learn the system with adaptive step solvers lead to numerical errors, while fixed-step size solvers struggle to accurately integrate as the dimensions operate on very different scales. The results of RK4 and our method in learning the ROBER system are shown in \cref{ROBER_comp}. Note that the y-axis of the plots are in log-scale. The trajectory obtained from RK4 can match the dimension that operates in the $10^{4}$ scale, but cannot concurrently adequately account for the trends in the other dimensions, which operate in much smaller scales. In our method, we observe the learned base ODE has dimensions that are much closer in scale, allowing for easier integration with large steps. The integrated base trajectories can then be passed through diffeomorphism $F$ to recover the integrated target trajectories. 
\Cref{Results_hard} (right) provides the performance and integration times of learning with our method and baseline numerical integrators, where we see that our method is more accurate than competing approaches while also requiring smaller execution times.


\section{Conclusions, Limitations and Societal Impacts}
We have proposed a novel approach to learning ODEs with neural networks, speeding up integration times by up to two orders of magnitude. Our method uses invertible neural networks to learn a diffeomorphism relating a desired target ODE to a \emph{base} ODE that is easier to integrate. We have shown that the base ODE can be linear or parameterised by a neural network. 
We have leveraged  the closed form solution of linear ODEs to provide remarkable speed-ups and allow constraints to be applied on the asymptotic properties of the learned ODE. 
We have also shown that, by using a base ODE parameterised by a neural network, we can learn ``difficult'' ODEs, with smaller networks to model dynamics. We have validated our approach by learning ODEs on synthetic and real-world data, and within continuous-depth neural network models. 

\textbf{Limitations:} We are restricted to learning ODE systems that evolve on some manifold that is diffeomorphic globally to Euclidean space (this includes Euclidean space). For example, we cannot learn an ODE that evolves on a torus. Standard ODE integrators are typically restricted to Euclidean space, so our method is still applicable to scenarios when differentiable numerical integrators are used. However, recent works have extended differentiable integrators to ODEs on more general manifolds \cite{manifold1,falorsi2020neural}. An extension of our approach to address this could be to segment the space up and use multiple diffeomorphisms. 

\textbf{Potential negative societal impact:} The learning of ODEs appears in many disciplines, including fields that guide decisions around social policy. Potential negative impacts can be mitigated by ensuring that results from models are validated by human judgement.

\bibliographystyle{abbrv}
\bibliography{main.bbl}

\newpage
\appendix
\section{Additional Results: Training times}
We present the training times for directly learning ODEs with our method, using a linear base ODE. These include the training times on the 3D Lotka-Volterra, and the robot imitation datasets, outlined in Sections 5.1.1 and 5.1.2 of the main paper. We run training for 1000 iterations, where in each iteration the batch includes the entire training set.
We see that, by leveraging the closed-form solution of linear ODEs, our method is able to also drastically speed up training. Additionally the parallel nature of passing through the invertible neural network allows more consistent training times across datasets.

\begin{table}[H]
\resizebox{\textwidth}{!}{%
\begin{tabular}{lcccccccc}
\hline
\multicolumn{1}{c}{} & \multicolumn{2}{c}{3D Lotka-Volterra} & \multicolumn{2}{c}{Imitation S} & \multicolumn{2}{c}{Imitation cube pick} & \multicolumn{2}{c}{Imitation C} \\ \cline{2-9} 
\multicolumn{1}{c}{} & Per iter (s) & Total (s) & Per iter (s) & Total (s) & Per iter (s) & Total (s) & Per iter (s) & Total (s) \\ \hline
Ours & 0.030± 0.002 & 29.57 & 0.029± 0.002 & 29.35 & 0.031± 0.007 & 31.34 & 0.030± 0.004 & 29.78 \\
Euler &  0.131±0.003 & 130.94 & 1.740± 0.017 & 1740.43 & 1.713± 0.016 & 1712.53 & 1.704± 0.010 & 1703.85 \\
Midpoint & 0.235±0.006 & 235.07 & 3.228± 0.027 & 3227.51 & 3.177± 0.028 & 3177.37 & 3.207± 0.025 & 3206.92 \\
RK4 & 0.469±0.005 & 468.52 & 6.671± 0.048 & 6671.44 & 6.388± 0.046 & 6388.10 & 6.441± 0.057 & 6440.61 \\
Dopri5 & 0.408±0.037 & 408.36 & 1.413± 0.034 & 1413.34 & 1.246± 0.023 & 1245.65 & 1.247± 0.022 & 1246.67 \\ \hline
\end{tabular}
}\caption{The training times in seconds with standard deviations, for 1000 iterations. By leveraging the closed-form solution of linear ODEs, training time with our method is consistently orders of magnitude faster than by using a differentiable numerical integrator.}
\end{table}

\section{Additional Results: Ablation Study}
We study the effects of the number of layers in the invertible neural network and number of parameters in the sub-network, which are the main hyper-parameters of the invertible neural networks used. To this end, we conduct ablation studies of the speed and performance of our method on the real-world datasets outlined in section 5.1.2 of the paper. Our basic model uses an invertible neural network with $5$ layers and sub-networks in the invertible network had $1500$ hidden dimension size. We alter the number of layers to be: $2$, $3$, $4$, $5$, $6$, $7$, $8$, and hidden dimensions of the sub-networks within the invertible network to be: $500$, $1000$, $1500$, $2000$, $2500$. The results are presented below:  

\begin{table}[H]
\resizebox{\textwidth}{!}{%
\begin{tabular}{cccccccc}
\hline
 &  & \multicolumn{2}{c}{Imitation S} & \multicolumn{2}{c}{Imitation cube pick} & \multicolumn{2}{c}{Imitation C} \\ \cline{3-8} 
No. Layers & Sub-Net Hid. Dim. Size & Int. time (ms) & MSE & \multicolumn{1}{c}{Int. time (ms)} & \multicolumn{1}{c}{MSE} & \multicolumn{1}{c}{Int. time (ms)} & \multicolumn{1}{c}{MSE} \\ \hline
2 & 1500 & 3.551± 0.585 & 122.40 & 2.993±0.103 & 41.51 & 2.914±0.049 & 20.69 \\
3 & 1500 & 4.589± 1.193 & 130.49 & 4.026±0.129 & 15.00 & 5.150±1.407 & 26.20 \\
4 & 1500 & 5.418± 0.746 & 24.54 & 6.294±0.939 & 26.18 & 5.374±0.382 & 10.33 \\
5 & 1500 & 6.461± 0.686 & 4.40 & 7.401±1.531 & 26.56 & 7.463±1.929 & 13.16 \\
6 & 1500 & 7.529± 0.698 & 8.17 & 8.993±2.212 & 17.16 & 8.994±2.781 & 18.27 \\
7 & 1500 & 9.426± 1.664 & 4.91 & 9.858±1.828 & 20.39 & 9.669±1.240 & 25.76 \\
8 & 1500 & 10.636± 2.541 & 5.62 & 10.732±2.111 & 14.56 & 10.958±2.518 & 6.57 \\
5 & 500 & 7.159± 1.475 & 5.62 & 8.018±2.109 & 19.37 & 7.315±1.483 & 6.22 \\
5 & 1000 & 6.972± 1.247 & 6.04 & 6.376±0.203 & 11.02 & 7.297±1.311 & 6.37 \\
5 & 1500 & 7.031± 1.289 & 4.40 & 7.321±1.236 & 26.56 & 6.901±0.802 & 13.16 \\
5 & 2000 & 7.787± 1.363 & 6.23 & 7.443±1.457 & 14.05 & 7.776±2.514 & 9.48 \\
5 & 2500 & 7.208± 1.628 & 10.92 & 6.521±0.082 & 11.66 & 7.611±1.687 & 5.59 \\ \hline
\end{tabular}
}\caption{Ablation study results of different configurations for the invertible neural network model.}
\end{table}

We see that as we increase the number of invertible network layers, the integration times increase, while the hidden dimension size of the sub-networks within the invertible network does not visibly affect the integration times. Overall, the generalisation performance improves as the number of invertible layers are used, up to some number of layers. Beyond this number of layers, adding layers does not vary performance significantly. Additionally, the hidden dimension sizes, for the values tested do not greatly vary the generalisation performance. 
\section{Proofs}
Proofs for Propositions 4.1 and 4.2 can be found in 
\cite{LeeManifolds} as Propositions 8.19 and 9.6.
\begin{customthm}{4.1}
Suppose two ODEs $\bb{x}'(t)=g(\bb{x}(t))$, $\bb{y}'(t)=f(\bb{y}(t))$ are related via $\bb{y}(t)=F(\bb{x}(t))$, where $F$ is a diffeomorphism. If the former ODE is asymptotically stable with $n_{e}$ equilibrium points $\bb{x}^{*}_{1},\ldots,\bb{x}^{*}_{n_{e}}$, then the latter ODE is also asymptotically stable, with equilibrium points $F(\bb{x}^{*}_{1}),\ldots,F(\bb{x}^{*}_{n_{e}})$.
\end{customthm}
\begin{proof}
First we show $F(\bb{x}^{*}_{1}),\ldots,F(\bb{x}^{*}_{n_{e}})$ are equilibrium points of ODE $\bb{y}'(t)=f(\bb{y}(t))$. 
By $\bb{y}(t)=F(\bb{x}(t))$, we can write the time derivatives $\bb{y}'$ at $F(\bb{x})$ as
\begin{equation}
    \bb{y}'(t)=f(F(\bb{x}(t)))=\frac{\mathrm{d}F(\bb{x}(t))}{\mathrm{d}t}=J_{F}(\bb{x}(t))g(\bb{x}(t)), 
\end{equation}
where $J_{F(\bb{x}(t))}$ is the Jacobian of $F$. $F$ is a diffeomorphism and hence invertible over its domain. By the inverse function theorem \cite{impl}, the Jacobian $J_{F}(\bb{x}(t))$ is invertible, and furthermore, by the invertible matrix theorem \cite{Mat_ref}, it has a null-space containing only the zero vector. Therefore, $\bb{y}'(t)=f(F(\bb{x}(t)))=J_{F}(\bb{x}(t))g(\bb{x}(t))=0$ if and only if $g(\bb{x}(t))=0$. As $g(\bb{x}^{*}(t))=0$ for $\bb{x}^{*}\in\{\bb{x}_{1}^{*}\ldots \bb{x}_{n_{e}}^{*}\}$, then we also have $f(F(\bb{x}^{*}(t)))=0$, hence $\bb{y}^{*}\in\{F(\bb{x}^{*}_{1}),\ldots,F(\bb{x}^{*}_{n_{e}})\}$ gives equilibrium points for $\bb{y}'(t)=f(\bb{y}(t))$. 

We now show asymptotically stability of $\bb{y}'(t)=f(\bb{y}(t))$, by the existence of a \emph{Lyapunov function} \cite{Lefschetz1962StabilityBL}, $V_{\bb{y}}:\mathbb{R}^{n}\rightarrow\mathbb{R}$, where $n$ is the dimension of $\bb{y}$, such that $\frac{\partial {V}_{\bb{y}}(\bb{y})}{\partial t}<0$ for all $\bb{y}\in\mathbb{R}^{n}\setminus\{F(\bb{x}^{*}_{1}),\ldots,F(\bb{x}^{*}_{n_{e}})\}$, and $\frac{\partial {V}_{\bb{y}}(\bb{y^{*}})}{\partial t}=0$ for $\bb{y}^{*}\in\{F(\bb{x}^{*}_{1}),\ldots,F(\bb{x}^{*}_{n_{e}})\}$. We assume the candidate function to be $V_{\bb{y}}=V_{\bb{x}}(F^{-1}(\bb{y}))$, where $V_{\bb{x}}$ is a valid Lyapunov function of the asymptotically stable $\bb{x}'(t)=g(\bb{x}(t))$, with $\frac{\partial{V}_{\bb{x}}(\bb{x})}{\partial t}<0$ for $\bb{x}\in\mathbb{R}^{b}\setminus \{\bb{x}^{*}_{1},\ldots,\bb{x}^{*}_{n_{e}}\}$ and $\frac{\partial{V}_{\bb{x}}(\bb{x}^{*})}{\partial t}=0$ for $\bb{x}^{*}\in \{\bb{x}^{*}_{1},\ldots,\bb{x}^{*}_{n_{e}}\}$. Consider the time derivative of the candidate function:

\begin{align}
   \frac{\partial V_{\bb{y}}(\bb{y})}{\partial t}&=\frac{\partial V_{\bb{y}}}{\partial \bb{y}}\frac{\partial \bb{y}}{\partial t}=\frac{\partial V_{\bb{y}}}{\partial \bb{y}}f(\bb{y})\\
   &=\Big(\frac{\partial{V_{\bb{x}}}}{\partial \bb{x}}\frac{\partial F^{-1}}{\partial \bb{y}}\frac{\partial F}{\partial \bb{x}}g(\bb{x})\Big)_{\bb{x}=F^{-1}(\bb{y})}\\
   &=\Big(\frac{\partial{V_{\bb{x}}}}{\partial \bb{x}}J_{F}(\bb{x})^{-1} J_{F}(\bb{x}) g(\bb{x})\Big)_{\bb{x}=F^{-1}(\bb{y})} && \text{By the inverse function theorem \cite{impl},}\\
   &=\Big(\frac{\partial{V_{\bb{x}}}}{\partial \bb{x}} g(\bb{x})\Big)_{\bb{x}=F^{-1}(\bb{y})} =\Big(\frac{\partial V_{\bb{x}}(\bb{x})}{\partial t}\Big)_{\bb{x}=F^{-1}(\bb{y})}.
\end{align}
Therefore, our candidate $V_{\bb{y}}$ is a valid Lyapunov function for $\bb{y}'(t)=f(\bb{y}(t))$. Thus, the system $\bb{y}'(t)=f(\bb{y}(t))$ is asymptotically stable.
\end{proof}


\section{Additional Implementation details}

We run all of our experiments on a machine with an Intel i7-3770k 3.50GHz processor, 32GB RAM and an NVIDIA GTX1080 GPU, with 8GB vRAM. For all of our experiments, we use the optimiser ADAM with step-size $10^{-4}$, except  for the experiments in the Latent ODE, which where we use the standard set-up from the Latent-ODE repository \cite{LatentODE}. The dynamics models of compared ODEs have the architecture: Input->dense(Input dimensions, 150)->tanh()->dense(150,150)->tanh()->dense(150,150)->tanh()->dense(150,150)->tanh()->dense(150,150)->tanh()->dense(150,output dimensions)->output. Except for the Latent-ODE comparisons where settings from the original repository \cite{LatentODE} is used, and for the stiff system, we train for 500 iterations with step-size $10^{-4}$, and then train with step-size $10^{-6}$ for 4500 iterations. For all of the experiments, except latent ODE experiments where we follow the original set-up, we train for $5000$ iterations in total. 

For all the experiments where we directly learn a dynamical system, we use an invertible neural network with 5 invertible layers, and sub-networks with one hidden layer of 1500 units. For non-linear base ODEs parameterised with a simple neural networks, we use the architecture: Input->dense(Input dimensions,30)->tanh()->dense(30,30)->tanh()->dense(30,30)->tanh()->dense(30,Output dimensions)->Outputs. Additionally, all learned dynamics, both with ours and compared methods, excepted when adhering to the original Latent-ODE set-up, were augmented with the same number of additional zeros as original state dimensions, for example 3 dimensional systems were augmented to 6 dimensions.  

The Lotka-Volterra system used has the dynamics:
\begin{align}
    x'(t)&=x(t)(0.75-0.75y(t))\\
    y'(t)&=y(t)(-0.75+0.75x(t)-0.75z(t))\\
    z'(t)&=z(t)(-0.75+0.75y(t))
\end{align}
for $t\in[0,7]$ with initial conditions $\{(5,5,1),(2,6,6),(3,1,4),(7,1,2),(6,2,4),\\(3,3,1),(2,2,2),(4,4,3),(3,3,4),(1,1,5)\}$.

The Lorenz system used has the dynamics:
\begin{align}
    x'(t)&=10(y(t)-x(t))\\
    y'(t)&=x(t)(28-y(t))-x(t)\\
    z'(t)&=x(t)y(t)-\frac{8}{3}z(t)
\end{align}
for $t\in[0,2]$ with the initial conditions $(0.15,0.15,0.15)$.

The Robertson's system used has the dynamics:
\begin{align}
    x'(t)&=-0.04x(t)+3\times 10^{4}y(t)z(t)\\
    y'(t)&=0.04x(t)-3\times10^{4}y(t)^{2}-10^{4}y(t)z(t)\\
    z'(t)&=3\times10^{4}y(t)^{2}
\end{align}
for $t\in[0,120]$ with the initial conditions $(1,0,0)$.

In the latent-ODE problem setup, an observable time-series is assumed to have latent variables which follow some ODE dynamics, and uses an \texttt{Encoder -> ODE -> Decoder} architecture where an ODE is used to model the hidden latent dimensions between the Encoder and Decoder. Note that a valid ODE is not guaranteed in the space of observable data, but only in the latent dimensions. Our set-up follows the repository given by \cite{LatentODE}, with the training settings for the Encoder and Decoder architecture as below:

\textbf{Periodic 100}: We train the entire model for 500 epochs with Adamax optimiser and an initial learning rate of $10^{-2}$. We sub-sample 5\% of the original time points and the size of the latent state is 10. The noise weight is set as 0.01 and the total number of time points is 100. For the Neural ODE architectures, there is one layer in the recognition ODE and one layer in the generative ODE, and 100 unit per layers. For the GRU unit there exists 100 units per layer for the GRU update network. All the above settings are exactly the same as the configuration given in repository \cite{LatentODE}.

\textbf{Periodic 1000}: Settings are the same as \emph{Periodic 100}, except that the total number of time points is set as 1000 to predict for finer time steps.

\textbf{Human Activity}: The model is trained for 200 epochs, with a dimensionality of 15 in the latent state. There are 4 layers in the recognition ODE and 2 layers in the generative ODE, and 500 units per layer. The GRU unit has exists 50 units per layer. These settings are exactly the same as the configuration given in the original repository \cite{LatentODE}.

\textbf{ECG}: Settings are the same as the classification task of \emph{Human Activity}, except that we use the ECG Heartbeat data available at \cite{ECG}.

\section{Additional Figures}
We provide figures for learning an additional Lorenz system for $t\in[0,5]$, with trajectory at initial condition $(-3.1,-1.15,8.15)$. We see that our method, with a base ODE parameterised by a neural network, can generate trajectories that closely match the ground truth:
\begin{figure}[h]
    \centering
    \includegraphics[width=\textwidth]{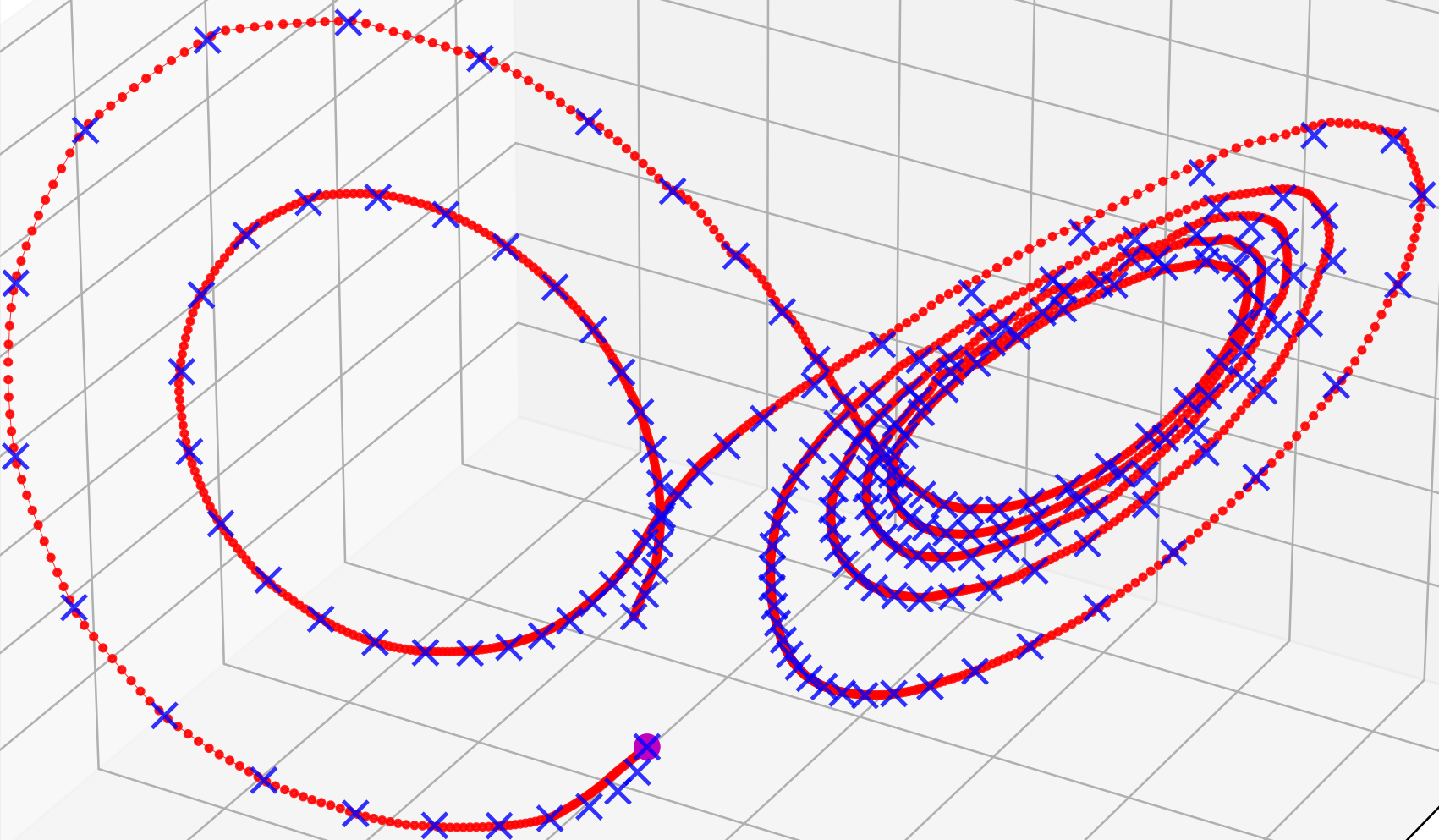}
    \caption{A learned Lorenz system with the generated trajectory, at 10x data resolution, and ground truth.}
    \label{fig:Lorenz1}
\end{figure}
\begin{figure}[H]
    \centering
    \includegraphics[width=\textwidth]{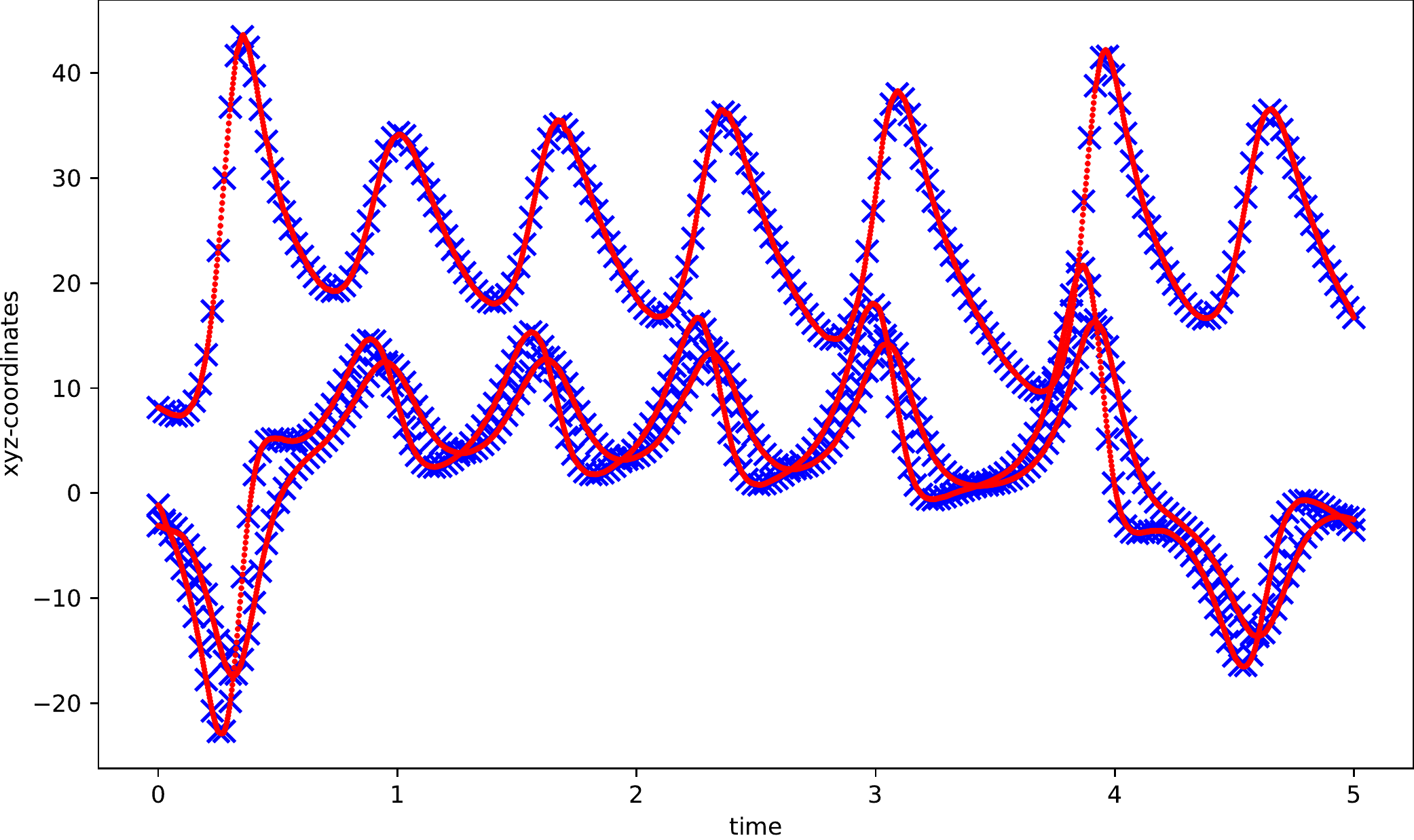}
    \caption{A learned Lorenz system with the generated trajectory, at 10x data resolution, and ground truth, rolled out in time}
    \label{fig:Lorenz2}
\end{figure}

We provide the change in coordinates over time, for the trajectory shown in figure 5(a) in the paper:
\begin{figure}[H]
    \centering
    \includegraphics[width=0.8\textwidth]{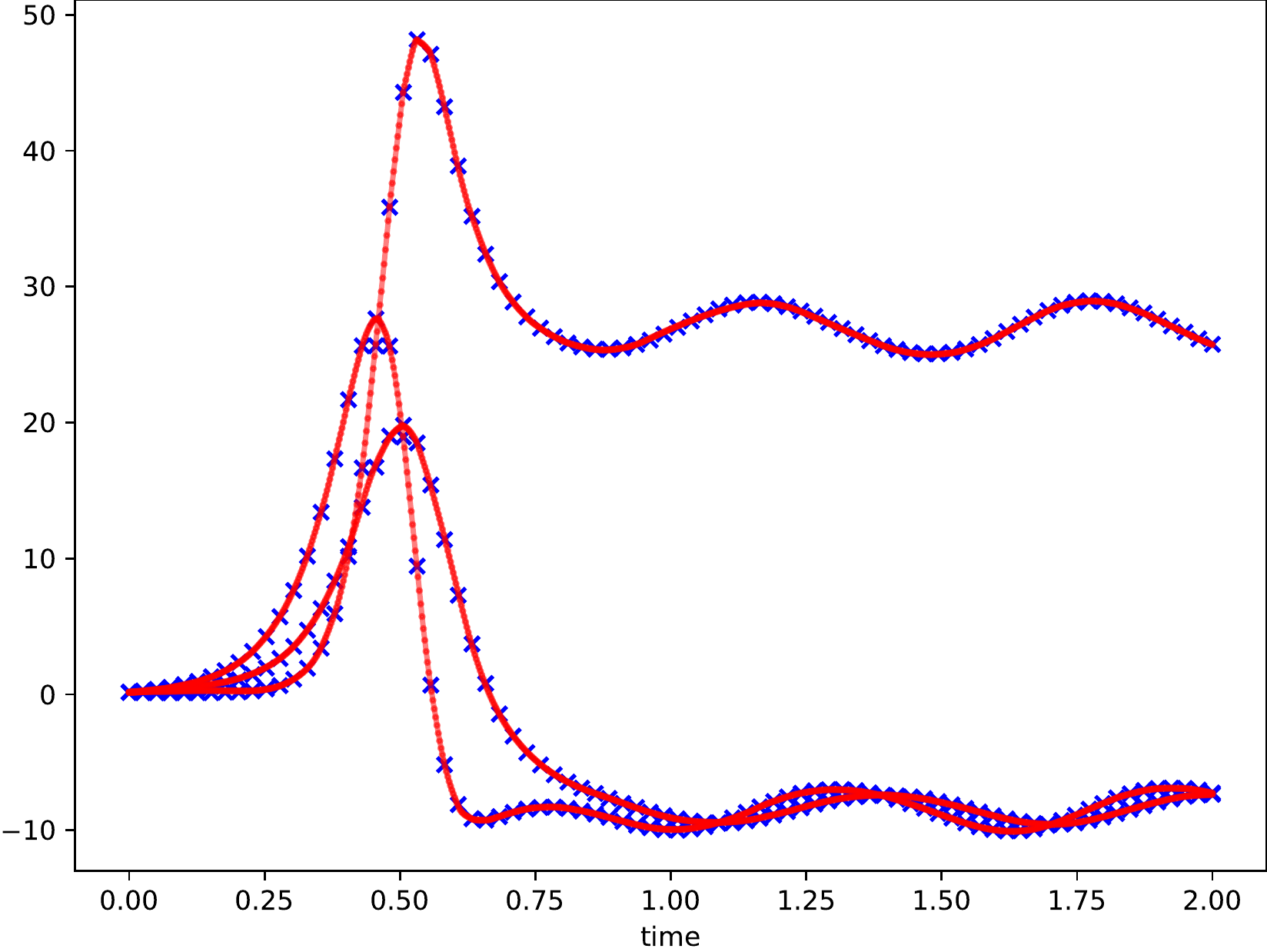}
    \caption{The corresponding plot showing the coordinates over the time interval of a learned Lorenz system over $t\in[0,2]$, at 10x data resolution, which corresponds to the 3d figure shown as fig 5(a) in the main paper.}
    \label{fig:my_label}
\end{figure}

The following figures show how the diffeomorphism finds a mapping between a stiff ODE with variations in very different scales across  dimensions, to a more manageable base ODE where the scales are more similar. The left plot is in log-scale, while the right plot shows the same ODEs in linear scale: 
\begin{figure}[H]
    \centering
    \includegraphics[width=\textwidth]{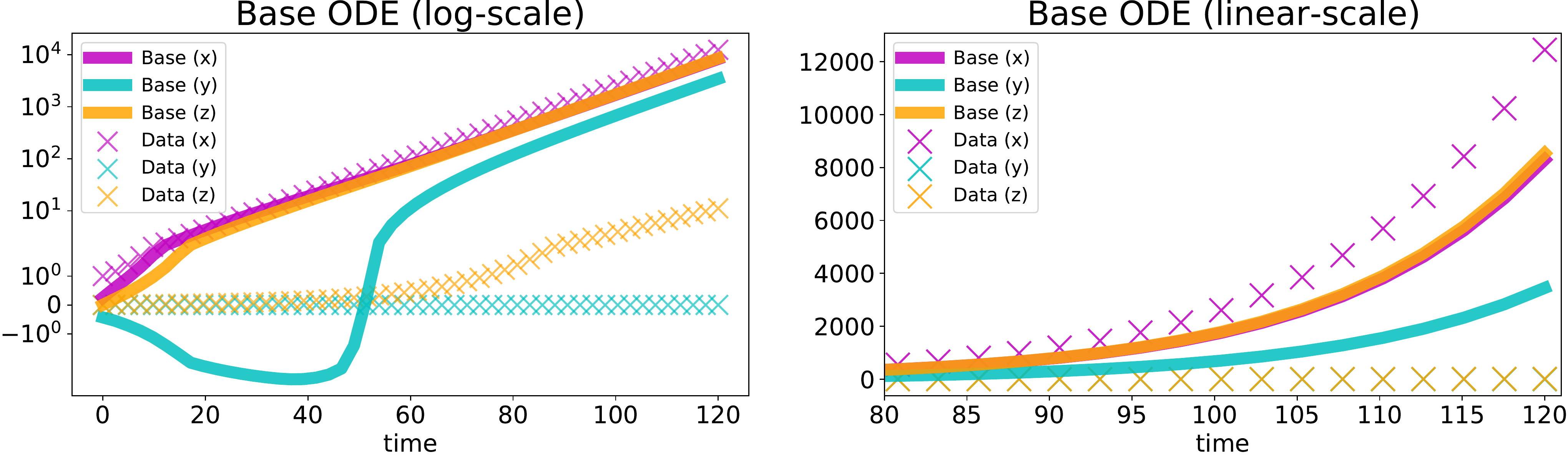}
    \caption{The data is generated from a stiff ODE 
    with large differences in scales (indicated in crosses) over the different dimensions. We see that the corresponding base ODE (indicated in solid lines) has changes with much more similar scales.}
    \label{fig:Stiff}
\end{figure}

We provide additional plots of trajectories, at different start points, from a learned Lotka-Volterra system. The ground truth data is in blue, while generated trajectory, of $10$x data resolution, is in red.
\begin{figure}[H]
\centering
\begin{subfigure}{0.46\textwidth}
    \centering
    \includegraphics[width=\textwidth]{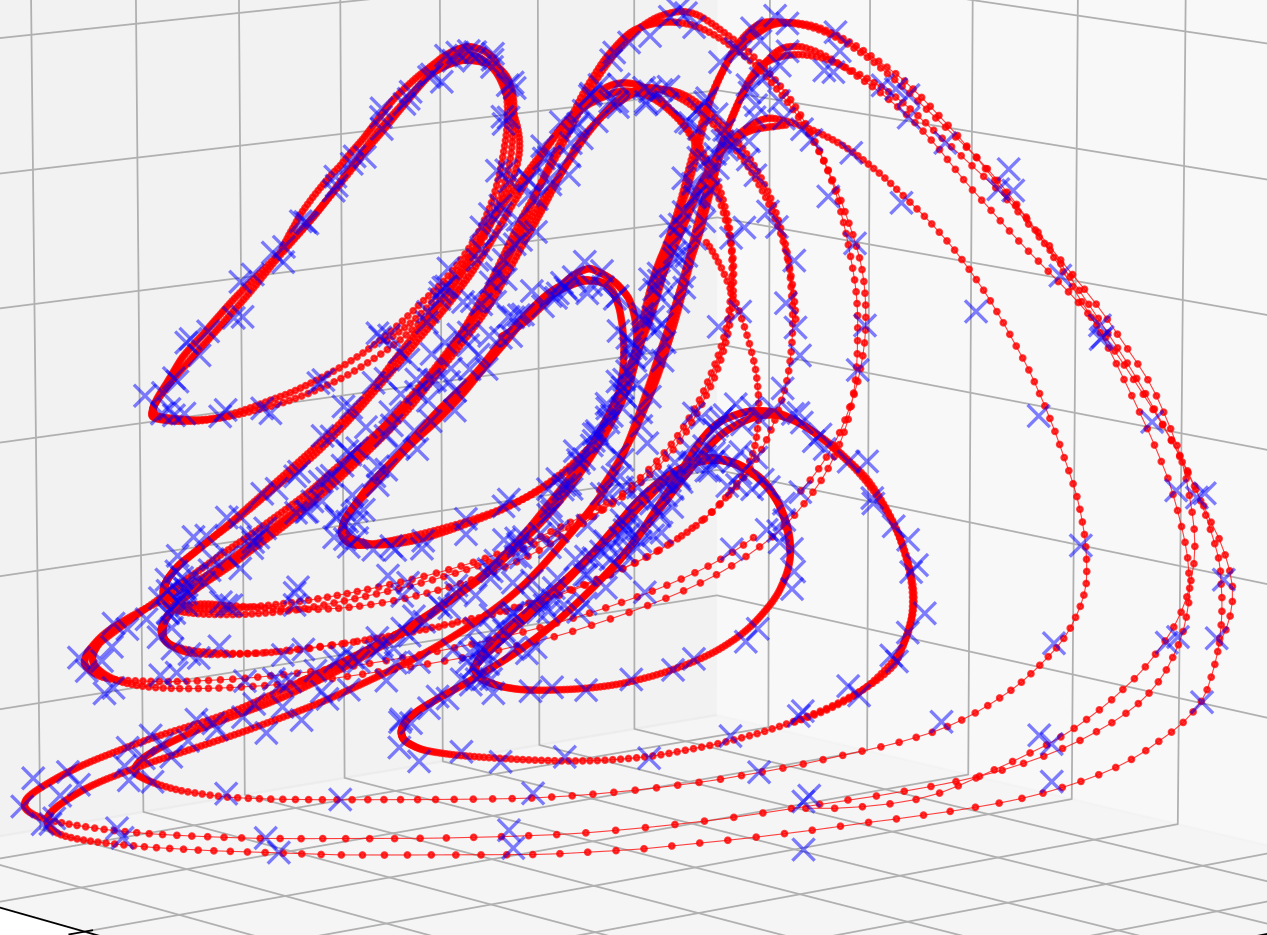}

\end{subfigure}
\begin{subfigure}{0.47\textwidth}
    \centering
    \includegraphics[width=\textwidth]{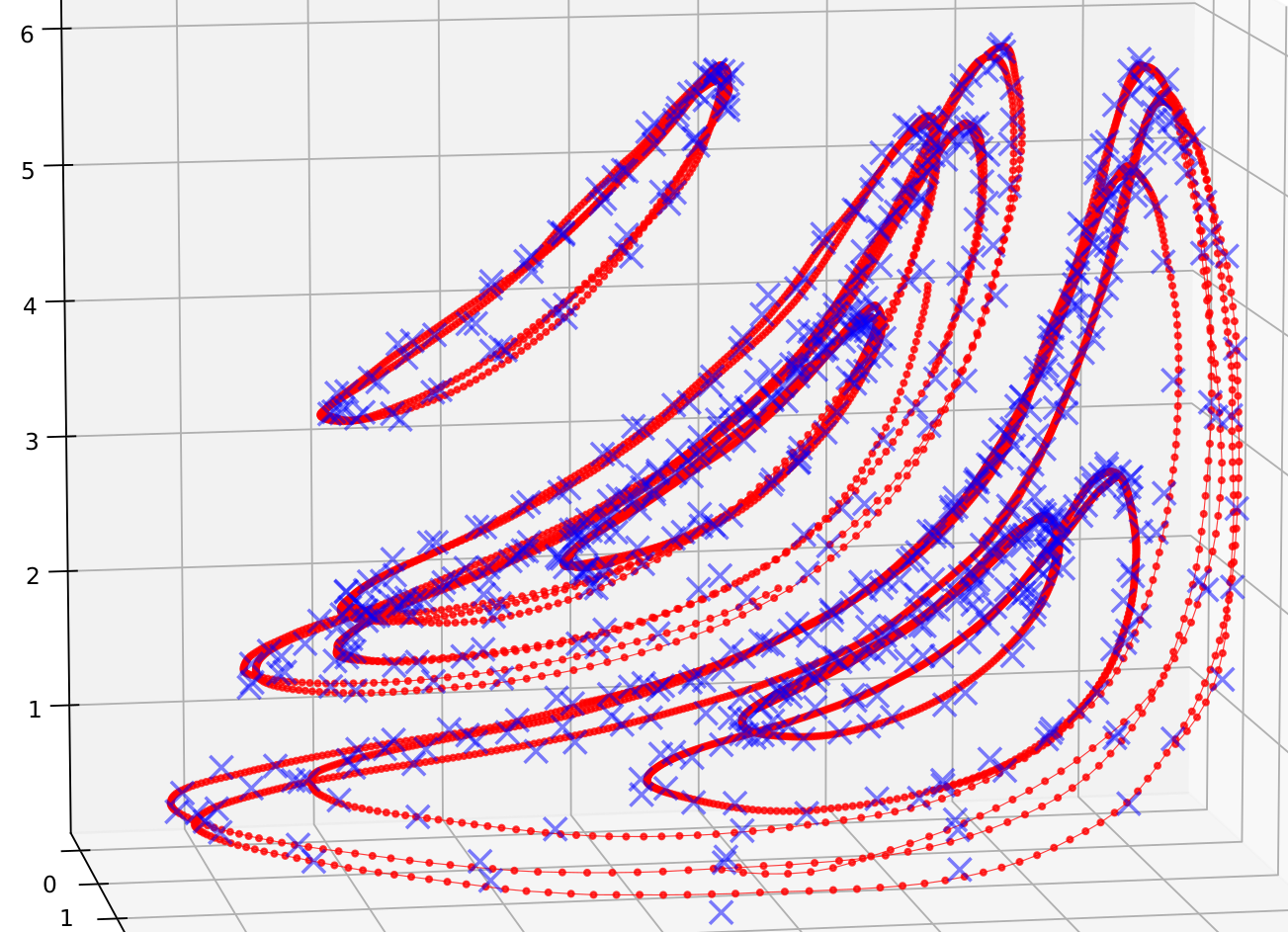}

\end{subfigure}
\caption{We see that trajectories from the learned Lotka-Volterra system, in red, closely matches the ground truth, in blue.}
\end{figure}

We provide an additional figure for trajectories generated at unseen starting points after being trained on the ``imitation C'' training data. The four generated trajectories are in red, while the ground truths are in blue. Our generated trajectories match the ground truth, and accurately capture the motion of drawing a ``C'' character.
\begin{figure}[H]
    \centering
    \includegraphics[width=0.45\textwidth]{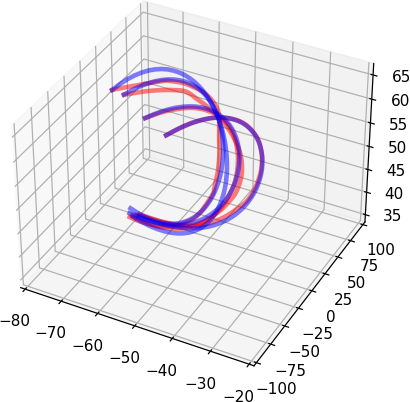}
    \caption{Robot motion trajectories in red, that imitate drawing a ``C'' character. The ground truth is given in blue.}
    \label{imitate-C}
\end{figure}

\section{Licenses for Packages}
Common scientific packages used in our code include: (i) Numpy \cite{harris2020array} (BSD license), for general linear algebra and miscellaneous math operations (ii) Matplotlib \cite{matplotlib_cite} (BSD compatible custom license), for plotting figures.

More specialised packages used include (i) {FrEIA} \cite{Freia} (MIT license), for invertible neural networks; (ii) {TorchDiffEq} \cite{torchdiffeq} (MIT license), for differentiable numerical integrators; (iii) {Pytorch} \cite{torch} (BSD license), for optimisation and automatic differentiation; (iv) Latent-ODE \cite{LatentODE} (MIT license), for latent ODE implementation.

\end{document}